\documentclass[letterpaper, 10 pt, conference]{ieeeconf}  

\IEEEoverridecommandlockouts                              

\usepackage{times}
\usepackage{epsfig}
\usepackage{graphicx}
\usepackage{amsmath}
\usepackage{amssymb}
\usepackage{booktabs}
\usepackage{comment}
\usepackage{subcaption}
\usepackage{multicol}
\usepackage{multirow}
\usepackage{floatrow}
\usepackage[font={small}]{caption}

\title{\LARGE \bf
Energy-Based Models for Cross-Modal Localization using Convolutional Transformers
}

\author{Alan Wu$^{1}$ and Michael S. Ryoo$^{2}$
\thanks{$^{1}$Alan Wu is with MIT Lincoln Laboratory, Lexington, MA, USA
        {\tt\small Alan.Wu@ll.mit.edu}}%
\thanks{$^{2}$Michael S. Ryoo is with the Department of Computer Science, Stony Brook University,
        Stony Brook, NY, USA
        {\tt\small mryoo@cs.stonybrook.edu}}%
\thanks{Distribution A. Approved for public release; distribution unlimited. OPSEC Number: 6834. This material is based upon work supported by the Department of the Army under Air Force Contract No. FA8702-15-D-0001. Any opinions, findings, conclusions or recommendations expressed in this material are those of the author(s) and do not necessarily reflect the views of the Department of the Army.}
}

\begin{document}

\maketitle
\thispagestyle{plain}
\pagestyle{plain}

\begin{abstract}

We present a novel framework using Energy-Based Models (EBMs) for localizing a ground vehicle mounted with a range sensor against satellite imagery in the absence of GPS. Lidar sensors have become ubiquitous on autonomous vehicles for describing its surrounding environment. Map priors are typically built using the same sensor modality for localization purposes. However, these map building endeavors using range sensors are often expensive and time-consuming. Alternatively, we leverage the use of satellite images as map priors, which are widely available, easily accessible, and provide comprehensive coverage. We propose a method using convolutional transformers that performs accurate metric-level localization in a cross-modal manner, which is challenging due to the drastic difference in appearance between the sparse range sensor readings and the rich satellite imagery. We train our model end-to-end and demonstrate our approach achieving higher accuracy than the state-of-the-art on KITTI, Pandaset, and a custom dataset. 


\end{abstract}

\section{Introduction}\label{introduction}

The ability to perform accurate localization within a given environment is key to vehicle navigation. Range sensors and RGB cameras have been employed on autonomous vehicles to describe its surroundings and help determine its pose on a map prior. With the growing prevalence of lidar sensors on vehicles due to its increasing affordability and its advantage over cameras in poor lighting and adverse weather conditions, lidar-based maps are often used for localization in the absence of GPS. As the immediate environment of the vehicle is described in the form of point clouds, this representation of the vehicle's state has been used in conjunction with maps built using onboard sensors of the same modality to perform localization in both 2D and 3D spaces \cite{chen2020survey}.  However, maps derived from range sensors are not available for many regions of the world, as they are often time-consuming to collect and expensive to compose, often leading to limited area coverage and constrained access. As a compelling alternative that is widely available and easily accessible, satellite imagery covers more regions of the world and contains rich structural information that can be correlated with lidar data.

\begin{figure}
{\includegraphics[width=8cm]{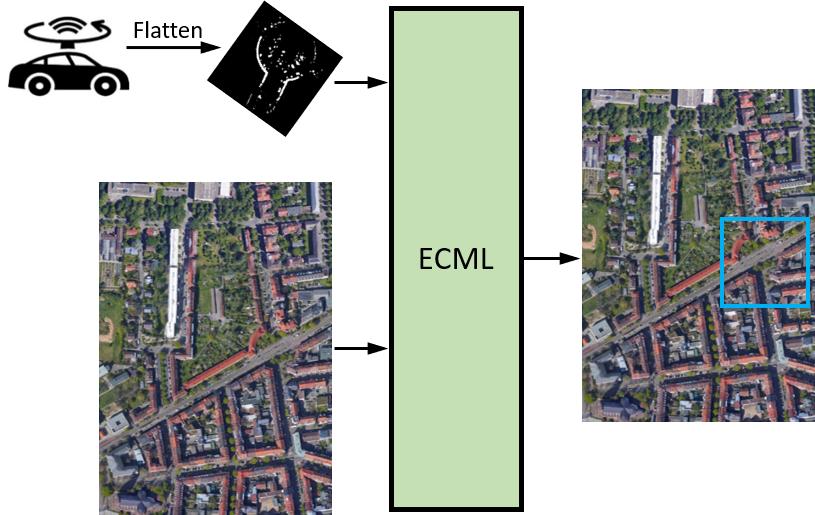}
\caption{Overview of our system: Energy-based Cross-Modal Localization (ECML). Vehicle is localized against satellite imagery via lidar point clouds flattened into rotated birds-eye view (BEV) images. Candidate satellite tiles are extracted from the map and paired with the BEV lidar image to find pose similarity: lidar-satellite pairs with high similarity achieve low energy. Here, the vehicle is located at the center of the blue box.}\label{fig:SystemDiagram}}
\end{figure}

We address the problem of localization across modalities at the metric (i.e. pixel) level by training an Energy-Based Model (EBM) \cite{LeCun06atutorial} to learn a similarity measure between the richly textural RGB satellite images and the sparse ground-based range sensor data (e.g., lidar). To better correlate with the structural information contained in the overhead imagery, we convert online lidar readings to birds-eye view (BEV) lidar images as done in \cite{barsan}, and later in \cite{tang-rslnet} and \cite{tang-self}. The energy function maps a lidar-satellite image pair to a scalar energy that reflects the degree of similarity between their poses. This simple yet flexible framework suits our localization task particularly well in scaling to map priors of arbitrary sizes and regions of coverage. 

As the backbone of EBMs can take on various forms, we leverage the recent advancements of convolutional neural networks (CNNs) and transformers for our task. The advent of the transformer architecture initially developed for natural language processing (NLP) applications \cite{vaswani2017} has greatly impacted the computer vision landscape upon the introduction of the visual transformer (ViT) \cite{vit2020} as it has outperformed traditional CNN architectures in the image classification task. Many works \cite{vit_survey}\cite{vit_survey2} have shown that purely transformer-based architectures achieved higher accuracy than their CNN counterparts, but required significantly more data to train. The authors of ViT postulated that the lack of inductive bias in transformers but inherent to CNNs have put it at a disadvantage when there is insufficient data. Some works \cite{cct2021}\cite{ceit} have had success taking advantage of the benefits of both architectures by combining elements of each to form convolutional transformers (CT) to achieve high accuracy but with less data than ViT. 

We follow the method pursued by \cite{cct2021}. Instead of dividing the input image into patches and tokenizing the image directly as in ViT, the image is first fed into convolutional layers before tokenization so that we can take advantage of the inductive bias property of CNNs as well as the highly effective attention mechanism of the transformer encoder. This is important to our task of correlating lidar BEV and satellite images that relies on retaining structural information of input images that would otherwise have been lost at the boundaries of patches via direct image tokenization. 

The main contribution of this paper is a method to localize a lidar-mounted vehicle on a satellite map prior scaled to an arbitrary size in GPS-denied environments. Our model (Fig.~\ref{fig:SystemDiagram}), Energy-based Cross-Modal Localization (ECML), learns to minimize the energy between a lidar-satellite pair with high pose similarity. Our contribution also includes an approach to accelerate inference for large map priors and a custom dataset collected in desert regions of southeastern California, a sandy environment prone to dust storms and yielding conditions that are not typically present in publicly available datasets. We validate our approach on the widely used KITTI \cite{kitti} dataset, the more recent Pandaset \cite{pandaset} dataset, and our custom dataset.

\section{Related Work} \label{related work}

\subsection{Cross-modal Localization with Overhead Imagery}

Localization using overhead imagery is a well studied subject. Various sensors such as cameras \cite{noda2010camera_aerial}\cite{pink2008visualmatching}\cite{wolcott2014visual_lidarmaps}\cite{churchill2013longtermloc}, lidars \cite{xu2017cam_lidarmap}\cite{carle2010lidarorbital3D}\cite{kummerle2009prior}, and radars \cite{tang-self}\cite{barnes2020radar} have been used against pre-built map priors. A common approach shared by many works is to match features between a query image from the perspective of the vehicle and an aerial map image. However, much effort has gone into handcrafting the features for specific scenarios that do not generalize well to unseen cases. Some approaches \cite{leung2008groundair}\cite{li2014geolocation} extracted edge features of buildings and applied chamfer matching on geometric shapes and lines. Relying heavily on road lane markings, \cite{noda2010camera_aerial} extracted SURF features from known landmarks while \cite{pink2008visualmatching} used a labor intensive method based on Canny edge detection that required manual removal of false positives. 
More recently, various sensor modalities have been used to match against custom lidar intensity maps such as synthesized camera viewpoints \cite{wolcott2014visual_lidarmaps}, satellite images \cite{veronese2015aerial}, and camera depth \cite{xu2017cam_lidarmap}. We use readily available satellite maps that require minimal processing.






\subsection{Learning-based Localization Approaches}
Numerous works have applied learning-based approaches to solving the problem of outdoor localization with prior maps. These can be broadly divided into three categories: pose estimation from visual and range sensor odometry, general place recognition, and metric level localization. 


\subsubsection{Pose Estimation}

Among the earlier works to apply deep learning to localization, PoseNet \cite{kendall2016posenet} predicted camera pose from images showing scenes that had been previously seen during end-to-end training, but from a different perspective. This relocalization task was achieved from regressing pose directly from egocentric images. Several extensions have improved upon performance by modeling pose uncertainty \cite{kendall2016modelling}\cite{Cai2018AHP}, and introducing geometry-aware loss functions \cite{kendall2017geometric}. L3-Net \cite{lu2019l3net} used point cloud data to learn pose from local keypoint descriptors through PointNet \cite{qi2017pointnet} followed by 3D CNNs. More recently, \cite{ratz2020oneshot} extracted segments from lidar scans to learn segment-level descriptors to match geometric segments of maps stored in a database. The sharper geometry resulting from sparse segments augmented with image patch descriptors have yielded impressive performance. Sharper geometry from sparsity of segmented data is analogous to our BEV of lidar data, which makes the data more sparse, but yields better geometry. Similar to our approach, it used \textit{one-shot} localization, as opposed to requiring a sequence of inputs when implementing a particle filter, for example. However, these approaches did not use prior maps, but rather built the maps while collecting their data, whereas we leverage widely available satellite imagery.

\subsubsection{General Place Recognition and Geolocalization}


Localization through image geo-tagging is another popular line of research, where images are geo-tagged and then used as a reference for online images during inference. Some works \cite{workman2015widearea} and \cite{lin2015deepgeo} explored cross-view general place recognition by embedding representations of ground-level and aerial images using a Siamese network. Similarly, \cite{hu2018cvm-net} and \cite{hu2019imagebased} learned dense features for ground-to-air geo-localization. Wang et. al \cite{wang2020urban2vec} trained a triplet network to learn features from images and point-of-interest markers for various urban neighborhoods. Chu et al. \cite{chu2015accurate} built a feature dictionary and learned projection matrices to compute a pose probability distribution of the input camera view with respect to a satellite map prior. Several works used a particle filter localization framework to match ground and aerial images. FLAG \cite{wang2017flag} learned features from detecting vertical structures, \cite{viswanathan2014visualsatloc} used image descriptors from a warped BEV image, and \cite{kim2017siamese}\cite{downes2022siamese} trained a Siamese network to learn common features of ground-satellite pairs while \cite{zhu2022triplet} used a visual transformer with triplet loss. Implementations of particle filters, such as these works, often assume prior knowledge of traversable areas such as roads to limit their initial search to these areas in order to achieve convergence, whereas we do not assume any prior knowledge of the map contents.  In addition, these works did not localize to the metric level.

\subsubsection{Metric Localization Using Overhead Imagery}



There has been much interest recently in applying deep learning techniques to metric localization. Ma et al. \cite{ma2019semanticHDmaps} leveraged high-definition (HD) semantic maps from lidar point clouds to localize the vehicle on highways using lane markers and road signs. Barsan et al. \cite{barsan} instead used unannotated lidar intensity maps and \cite{wei2019binmaps} investigated using both types of HD maps, but more efficiently by compressing them into binary representations. All these works, however, localize against maps built using the same modality as their online sensor. Others \cite{tang-rslnet}\cite{tang-self} employed overhead imagery as their map prior while using either a radar or lidar onboard sensor to perform cross-modal localization. Similar to \cite{barsan} and \cite{lu2019l3net}, they searched exhaustively for the optimal pose offset instead of direct regression. Our approach differs in two main aspects: (1) we do not require GPS at inference and (2) our map is much greater in size than the online BEV image whereas theirs is of similar size but slightly offset due to a noisy GPS. While their modalities have significant overlap, our map prior can be orders of magnitude greater than the online image, leading to a much larger search space.


\begin{figure}
    \centering
    \includegraphics[width=\textwidth]{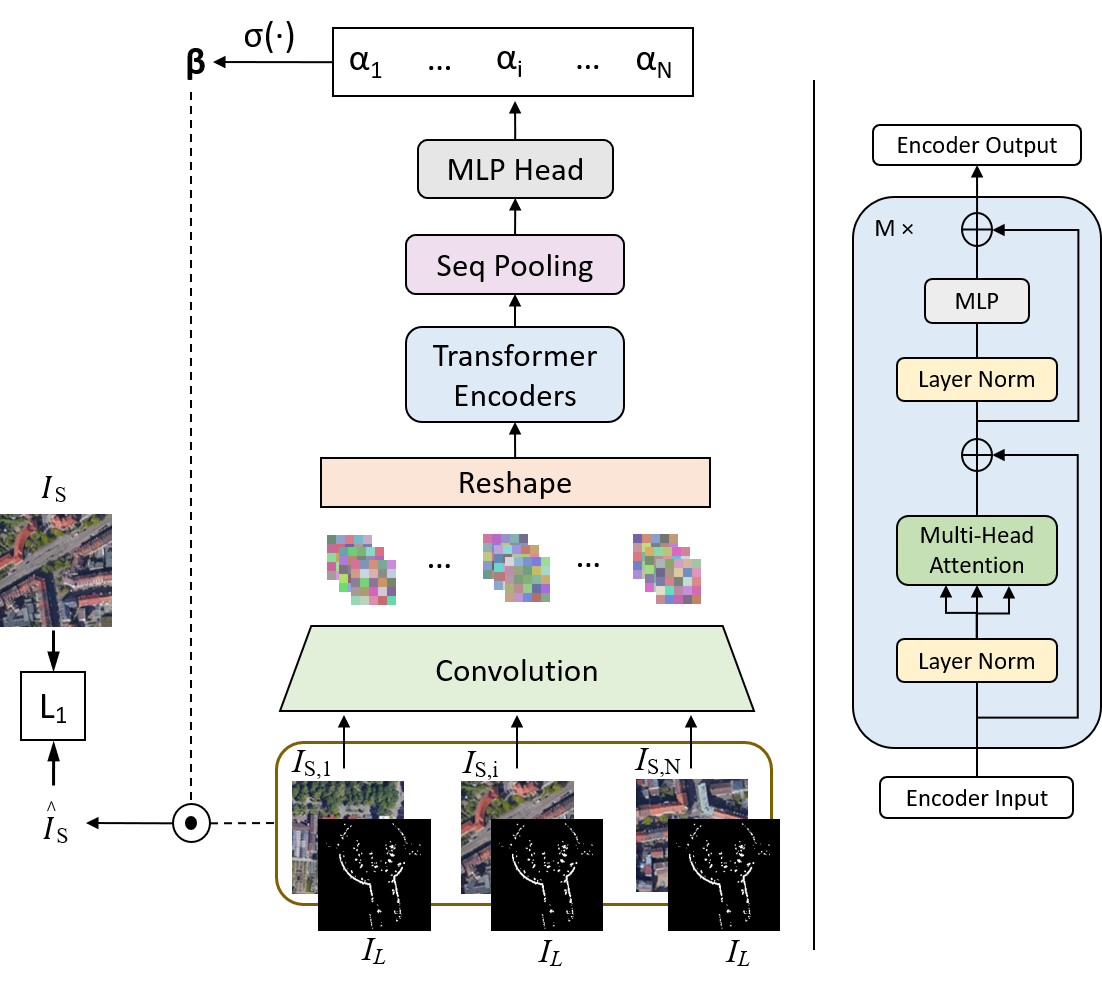}
    \caption{Our cross-modal localization network ECML-CT. Birds-eye view (BEV) lidar images concatenated ($\oplus$) with an array of candidate satellite tiles serve as input to the model. We construct $\hat{I}_S$ from the dot product $\pmb{\beta}\cdot\textbf{I}_S$ where the coefficients $\pmb{\beta}$ are the softmax output of similarity scores \textbf{A} and the satellite array $\textbf{I}_S$ is derived from $I_{S_{map}}$. At inference, we simply select the lidar-satellite pair that yields the highest similarity score $\alpha^*$ without using the softmax $(\sigma(\cdot))$ or the dot product $(\odot)$.}
    \label{fig:clect}
\end{figure}

\subsection{Convolutional Transformers}
The predominant architecture in NLP for sequence-to-sequence mapping, the transformer \cite{vaswani2017} inspired the visual versions that have outperformed CNN counterparts in image classification \cite{vit2020}, object detection \cite{detr}\cite{detr3d}, anomaly detection \cite{vt-adl}, semantic segmentation \cite{vt}\cite{point-transformer}, and autonomous driving \cite{transfuser} tasks. However, pure visual transformers require substantially more data than CNNs, partly because of their lack of inductive bias \cite{vit2020}\cite{cct2021}. Various works have attempted to alleviate the need for large amounts of data by using strong data augmentation and knowledge distillation \cite{deit} or by incorporating convolutional layers to extract the lower level features of the image before tokenizing the feature maps and using the transformer to learn higher level features \cite{vt}\cite{ceit}\cite{cct2021}. Our model builds upon \cite{cct2021} and 
is the first to leverage transformers for cross-modal localization.

\section{Methodology} \label{methodology}
\subsection{Energy-Based Models}
Energy-Based Models \cite{LeCun06atutorial} provide a simple yet powerful and flexible framework 
that relies on a scalar energy to capture dependencies between variables. Their effectiveness has been demonstrated in multiple domains including image generation \cite{xie2016}\cite{nijkamp2019}, classification \cite{grathwohl2019}, object detection \cite{gustafsson2020}, machine translation \cite{tu2020}, structured prediction \cite{rooshenas2019}\cite{gygli2017}, reinforcement learning \cite{du2020}, and continual learning \cite{cl_li}. 
To our best knowledge, cross-modal localization is a novel application for EBMs.


Energy-Based Models rely on the principle that any probability density $p(\mathbf{u})$ for $\mathbf{u} \in \mathbb{R}^K$ can be expressed as the Boltzmann distribution:
\begin{equation}\label{eq:ebm-general}
    p_\mathbf{w}(\mathbf{u}) = \frac{\exp(-E_\mathbf{w}(\mathbf{u}))}{Z(\mathbf{w})}~,~
    Z(\mathbf{w}) = \int_\mathbf{u}\exp(-E_\mathbf{w}(\mathbf{u}))
\end{equation}
where $E_\mathbf{w}(\mathbf{u}): \mathbb{R}^K \rightarrow \mathbb{R}$, known as the \emph{energy function}, maps each data point to a scalar energy value, and $Z(\mathbf{w})$ is the normalizing partition function. Thus, one can parameterize an EBM using any function that takes $\mathbf{u}$ as the input and returns a scalar. In our application, the energy function $E_\mathbf{w}(\mathbf{u})$ is a convolutional transformer parameterized by $\mathbf{w}$. 

\subsection{Energy-Based Models for Cross-Modal Localization}
We use lidar data to find vehicle pose within a satellite image. 
Point cloud data from range sensors have shown to be effective in localization tasks when a flattening step is applied to produce a birds-eye view (BEV) of the environment surrounding the vehicle. \cite{barsan}  We use this BEV representation to localize within a large map area.  While our primary objective is to estimate vehicle position in the x,y-plane, our method also solves for the rotational offset between the online lidar image and the satellite map prior.

Given a greyscale lidar image $I_L \in \mathbb{R}^{d\times{d}\times1}$ and an RGB satellite map $I_{S_{map}} \in \mathbb{R}^{D\times{D}\times3}$ ($D >> d$), we seek to find the $(x,y)$ pixel location of the vehicle on $I_{S_{map}}$. To accomplish this, we learn an energy function $E_\textbf{w}(I_{L}, I_{S_i})$ where $I_{S_i} \in {\mathbb{R}}^{d\times{d}\times3}$ is the $i$-th image cropped from $I_{S_{map}}$ at coordinate $(x_i,y_i)$. 
 Since there may be rotational offsets between $I_L$ and $I_{S_{map}}$, a set of $N_\theta$ rotated lidar images $I_{L}(\theta)$ serves as candidates to describe the state of the vehicle. Therefore, there is a total of $N_S = M_S \times N_\theta$ lidar-satellite image pairs, where $M_S$ is the number of satellite images cropped from $I_{S_{map}}$ via a sliding window. The pose of the vehicle is determined by the lowest energy achieved among all lidar-satellite pairs:
\begin{equation} \label{eq:pose}
    (x^*,y^*,\theta^*) = \underset{x,y,\theta \in \mathcal{X,Y},\Theta}{\arg\min} \;E_\mathbf{w}(I_{L}(\theta),I_S(x,y))
\end{equation}
\vspace{-0.2cm}

We adapt the general formulation of EBMs from Eq.~\ref{eq:ebm-general} for cross-modal localization. Given input $I_L$ and a discrete set $\mathcal{I}_S$ of satellite tiles derived from satellite map $I_{S_{map}}$, we use the Boltzmann distribution to define the conditional likelihood of tile $I_S$ given $I_L$ as follows:

\vspace{-0.3cm}
\begin{align}\label{eq:ebm-cml}
    &p_{\mathbf{w}}(I_S | I_L) = \frac{\text{exp}(- E_{\mathbf{w}}(I_L, I_S))} {Z(\mathbf{w}; I_L)},\\ 
    &\quad Z(\mathbf{w}; I_L) = \sum_{I'_S \in\mathcal{I}_S} \text{exp}(- E_{\mathbf{w}}(I_L, I'_S)) \nonumber 
\end{align}
\vspace{-0.1cm}where $E_{\mathbf{w}}(I_L, I_S): (\mathbb{R}^{U}, \mathbb{R}^{V}) \rightarrow \mathbb{R}$ is the energy function mapping a lidar-satellite pair $(I_L, I_S)$ to a scalar energy value, and $Z(\mathbf{w}; I_L)$ is the partition function for normalization. The objective of the EBM is to minimize energy at ground truth tile $I_S$ and increase the energy at other tiles $I'_S$.

The key component of our EBM is a convolutional transformer (CT) adopted from \cite{cct2021}. The CT architecture applies particularly well to our application as the early convolutional layers help better preserve local structural information that would otherwise have been lost by direct tokenization of input images (i.e. patch boundaries). The class and position embeddings used in the ViT architecture are eliminated in the CT approach. The implementation of the sequence pooling layer \cite{cct2021} renders such embeddings unnecessary, leading to a more compact transformer. In addition to an EBM with a CT backbone (i.e. ECML-CT), we also investigated using a CNN backbone (i.e. ECML-CNN), which entails 3 down-convolutional layers followed by six ResNet blocks \cite{he2015deep}, where the mean is taken along the height, width, and channel dimensions to attain the same network output $\textbf{A}$ as ECML-CT, as detailed in the next section.

\subsection{Training ECML-CT}\label{ecml-ct}
To estimate $E_\mathbf{w}(I_L, I_S)$, we train the convolutional transformer (CT) in a self-supervised manner to construct the predicted satellite tile $\hat{I}_S$ from a weighted ensemble of tile candidates. The model architecture for ECML-CT is shown in Fig.~\ref{fig:clect}. It takes as input a lidar-satellite pair ($I_{LS_{i}} = I_{L}\oplus I_{S_{i}}$) concatenated along the channel axis where $I_{LS_{i}} \in R^{H \times W \times C}$. We use two convolutional layers with ReLU activation and max pooling to obtain an intermediate representation $z \in R^{H_{0} \times W_0 \times p}$ so that $p$ is equivalent to the embedding dimension of the transformer backbone used in ViT. We reshape $z$ to $z_0\in R^{l\times p}$ where $l=H_{0} \times W_{0}$, which is the length of the embedding sequence. $z_{0}$ is then fed to the transformer encoder. The sequential output of the M-layer transformer encoder is followed by a sequence pooling step \cite{cct2021} where the importance weights are assigned for the output sequence, which is subsequently mapped to a similarity score for the lidar-satellite pair in the MLP head.

Each batch consists of physically proximal and distant lidar-satellite pairs with $N_S$ pairs in total, and thus yielding $\textbf{A}=[\alpha_1, \alpha_2, ... \alpha_{N_S}]^T\in{\mathbb{R}^{N_S}}$, which we use as similarity scores at inference. To minimize the energy, which is the negative output (-$\alpha_i$) of the CT, we aim to generate a satellite tile $\hat{I}_S$ from a weighted sum of the candidate satellite tiles $\mathbf{I}_{S}$ by applying a softmax operation to $\mathbf{A}$ to obtain a normalized probability distribution $\pmb{\beta}$ such that $\sum_{i=1}^{N_S} \beta_i = 1$. The elements of $\pmb{\beta}$ serve as coefficients for a linear combination of satellite candidates $\mathbf{I}_{S}$ in constructing the predicted satellite image $\hat{I}_S = \pmb{\beta} \cdot \textbf{I}_{S}$, with the aim that the $\beta_i$ for the vehicle location will be close to 1 while others will be close to 0. The likelihood of each candidate reflects the match quality between lidar-satellite image pairs, ($I_{L}, I_{S_i}$), where $I_{L}$ is the aligned online lidar image and $I_{S_i}$ is a satellite tile candidate derived from $I_{S_{map}}$. 
We apply $L_1$ loss to the predicted satellite image: $\mathcal{L}(\hat{I}_{S},I_S) = {||I_S - \hat{I}_{S}||}_1$, where $I_S$ is the true satellite tile centered at the vehicle location. We apply a learning rate of 0.0005, batch size of 2, and run for 500 epochs using the Adam \cite{kingma2014adam} optimizer.

\subsection{Vehicle Pose Inference}\label{sec:pose_inference}
Since CTs and CNNs are not equivariant to rotation, we provide $N_\theta$ rotated lidar BEV candidates $I_{L}(\theta)$ at inference. The lidar image with the optimal rotation $I_{L}(\theta^*)$ paired with the satellite image at the vehicle location $I_{S}(x^*,y^*)$ will provide the highest score $\alpha^*$ of the score vector \textbf{A}$\in {\mathbb{R}}^{N_S}$ where $N_S = M_S \times N_\theta$ and $M_S$ is the number of satellite images cropped from $I_{S_{map}}$ via a sliding window. Vehicle pose is then found using Eq.~\ref{eq:pose}. Note that 
we take the $argmax$ of the scores to get optimal pose estimate, and no longer use the softmax or the dot product. 

\begin{figure}[b]
    \includegraphics[width=0.58\linewidth]{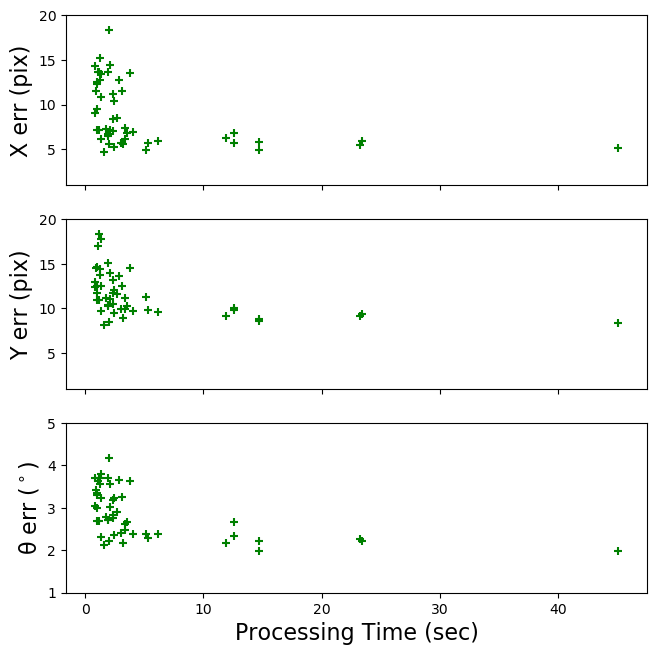}
    \caption{Pose errors for various values of $m$ and $n$ plotted against inference processing time, for map size of $351\times351$ m$^2$ and heading noise of $\pm10^{\circ}$.}
    \label{fig:accuracy_vs_time}
\end{figure}

As the map grows in size, sliding a window across every possible pixel would prohibit real-time inference. We propose a 2-stage inference approach that significantly decreases processing time. In the first stage, we compose a set of input pairs by skipping $m$ pixels in both x and y coordinates, and $n$ degrees, yielding $\tfrac{M_{S}}{m^2} * \tfrac{N_{\theta}}{n}$ pairs. We then take the top $k$ candidates with the highest similarity scores and perform a 2nd stage of inference where we sweep the area around those candidates in pixel and angle space in 1 pixel and $1^{\circ}$ increments, respectively. That is, we ``fill the gaps'' from the 1st stage around the top $k$ candidates by sweeping [$x_{j}-(m-1), x_{j}+(m-1)$], [$y_{j}-(m-1), y_{j}+(m-1)$], and [$\theta_{j}-(n-1), \theta_{j}+(n-1)$] where $j\in[1,k]$. The predicted pose is then derived from the lidar-satellite pair with the highest similarity score of the 2nd stage. We experimented with $m\in[1,8]$ and $n\in[1,4]$, and found $m=3$ and $n=2$ to result in an optimal tradeoff between pose accuracy and processing time. See Fig. \ref{fig:accuracy_vs_time}. Inference is performed in 1.59~sec on a Titan X GPU for an area of $351\times351$ m$^2$. 

\begin{figure}
  \frame{\includegraphics[width=.49\linewidth]{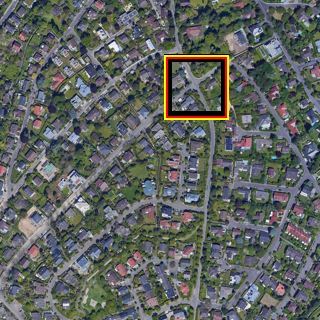}}
  \makebox[0pt][r]{
    \raisebox{0.1em}{
      \includegraphics[width=.1\linewidth]{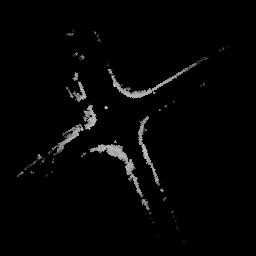}
    }\hspace*{9.2em}
    }
  \frame{\includegraphics[width=.49\linewidth]{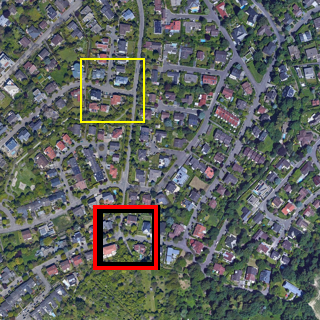}}
  \makebox[0pt][r]{
    \raisebox{0.1em}{
      \includegraphics[width=.1\linewidth]{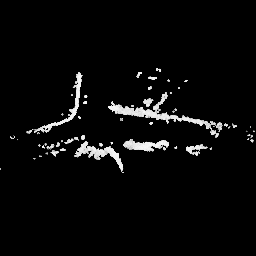}
    }\hspace*{9.17em}
    }
  \frame{\includegraphics[width=.49\linewidth]{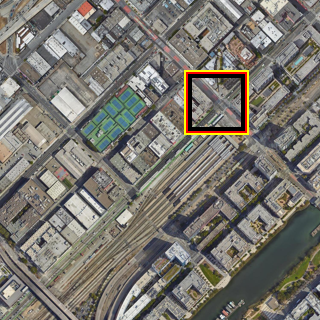}}
  \makebox[0pt][r]{
    \raisebox{0.1em}{
      \includegraphics[width=.1\linewidth]{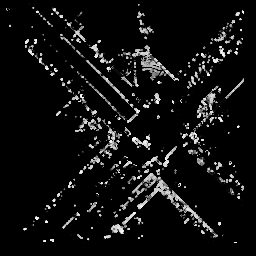}
    }\hspace*{9.2em}
    }
  \frame{\includegraphics[width=.49\linewidth]{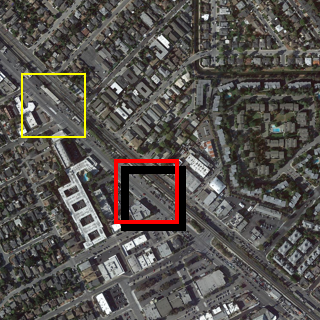}}
  \makebox[0pt][r]{
    \raisebox{0.1em}{
      \includegraphics[width=.1\linewidth]{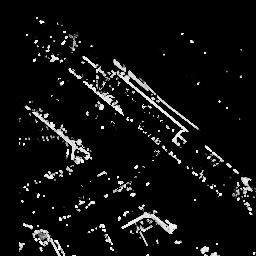}
    }\hspace*{9.17em}
    }
  \frame{\includegraphics[width=.49\linewidth]{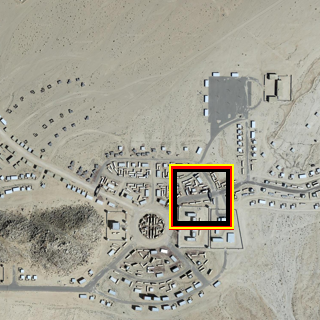}}
  \makebox[0pt][r]{
    \raisebox{0.1em}{
      \includegraphics[width=.1\linewidth]{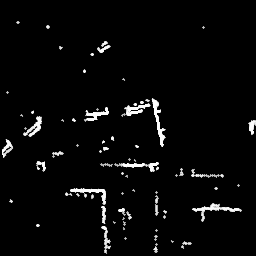}
    }\hspace*{9.2em}
    }
  \frame{\includegraphics[width=.49\linewidth]{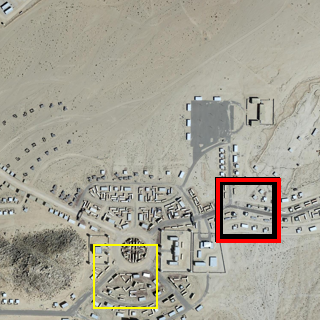}}
  \makebox[0pt][r]{
    \raisebox{0.1em}{
      \includegraphics[width=.1\linewidth]{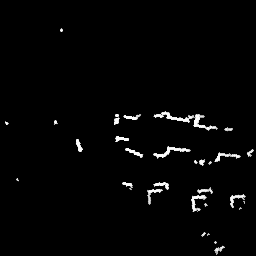}
    }\hspace*{9.17em}
    }
  \caption{Examples of online vehicle lidar image queries overlaid on the bottom-left corner of satellite map priors 
  and resulting localizations for KITTI (top), PandaSet (middle), and Custom (bottom) datasets. Black, red, and yellow boxes are ground truth, CT-predicted, and CNN-predicted locations, respectively. The maps are retrieved from a database so that the vehicle location is randomly positioned on the map to make the task more challenging and allow for \textit{one-shot} localization. The right column shows examples of failures by ECML-CNN but successes by ECML-CT. Note that the correct and incorrect predictions are very similar in appearance.}
  \label{fig:query_map}
\end{figure}

\section{Experiments and Results} \label{Experiments and Results}
\subsection{Datasets and Experimental Setup} 
We use the publicly available KITTI \cite{kitti} and Pandaset \cite{pandaset} datasets, and a custom dataset to validate our approach. While KITTI is a well-known and popular dataset collected in Karlsruhe consisting of mainly suburbs, the more recently composed Pandaset is set in San Francisco, where urban canyons are more prevalent and GPS is often denied. Our custom dataset is collected on and off roads in two desert towns in southeastern California.

For KITTI, we use 5 routes for training and 2 routes for test, with the split roughly at 16000 images and 1600 images, respectively. During inference, we use every 16th frame, which yields 100 waypoints along the routes, with each waypoint consisting of a lidar-satellite pair. For Pandaset, which is organized by route segments (i.e. subroutes), we use 93 subroutes for training and 10 subroutes for test; we follow a similar split ratio (7440 training images and 800 test images). For our custom dataset, we split our data in two different ways. In the first scenario (custom A), we combine data from both towns and use 205 subroutes for training and 47 subroutes for test, which amounts to a split of about 58500 images and 13500 images, respectively. In the second scenario (custom B), we use data collected in one town for training and the other town for test, which is more challenging since both the lidar data and the satellite imagery will be unseen by the model, as opposed to only the lidar data. The split for this scenario is 147 subroutes for training and 105 subroutes for test, which leads to about 42000 images and 30000 images, respectively. For all the datasets we extract 100 waypoints evenly from the test split to assess pose prediction accuracy. Odometry information used for heading and GPS coordinates used to retrieve satellite images for training were made available through onboard sensors for all datasets. No GPS data is used during inference. 

We convert lidar point clouds to birds-eye view (BEV) images to correlate with satellite imagery for vehicle localization. The sparse greyscale lidar BEV images, composed of only points registered above ground (i.e. $z {>} 0$), are $64{\times}64$ pixels with a resolution of 1.83 m/pixel, which is the same scale as all satellite imagery. Individual RGB satellite tiles derived from the satellite map are of the same image size and resolution. We vary the coverage area of the satellite map at each waypoint to help the model learn better for \textit{one-shot} localization, as adjacent vehicle waypoints have very different (x,y) pixel locations on maps $I_{S_{map,i}}$ and $I_{S_{map,i+1}}$.

To achieve metric-level localization, the position search space would span the entire satellite map minus the borders since all candidate satellite tiles would need to be wholly encapsulated within the map prior. Although an exhaustive search in the heading search space would span $[-180^{\circ},+180^{\circ}]$, we assume access to a coarse heading measurement, such as a noisy compass, is available at inference. 
Therefore, for our experiments, we add noise sampled uniformly from the range $[-10^{\circ},+10^{\circ}]$ to the ground truth heading $\theta_{gt}$ and propose candidates rotated in $1^{\circ}$ increments to cover the range of uncertainty. This results in the heading search space $\theta\in[(\theta_{gt}-10^{\circ}),(\theta_{gt}+10^{\circ})]$.


\subsection{Experimental Results}\label{experimental_results}
We compare our approach to other methods for metric localization as shown in Table \ref{tab:comparison}. The state of the art in cross-modal localization via live range sensing against satellite imagery that inspired our work is \cite{tang-rslnet}. However, different from our goal, theirs is to refine a coarse estimate from a noisy GPS fix, which implies that their search area of the satellite map is only slightly greater than the immediate online range sensor coverage area. That is, the size of their satellite map image is similar in size as the range sensor BEV image, whereas in our case the size of the satellite map image is much greater than the sensor BEV image. Having the same image sizes between the two modalities allows them to generate a synthetic image of the entire satellite map that takes on the appearance of a sensor BEV image. With both images now appearing in the same modality, the correlation surface is found by convolving the online BEV image with the synthetic BEV image, and the (x,y) position offset is found by taking the $argmax$ of the correlation matrix.

Our task is to localize the vehicle in GPS denied regions, and therefore the search space (i.e. map size) is much greater than the coverage area of the online lidar. With a satellite map size much greater than the range sensor BEV image size, we can no longer generate a range sensor-like synthetic image of the entire satellite map as a whole as done in \cite{tang-rslnet} and \cite{tang-self}. To adapt their approach to our task, we assemble all possible satellite tiles \textbf{I}$_S$ derived from the satellite map $I_{S_{map}}$ via a sliding window of size $d\times{d}$ and convert each tile to synthetic lidar images \textbf{I}$_{L_{gen}}$ using the image generator $g(\:\cdot\:;\mathbf{w_g})$ similar to that in \cite{tang-rslnet} with satellite tiles from our datasets as input: $g(\mathbf{I}_S;\mathbf{w_g}) \rightarrow \mathbf{I}_{L_{gen}}$ where $\mathbf{w_g}$ are learned parameters. Each generated lidar tile is then correlated with the rotated online lidar images \textbf{I}$_{L}(\theta)$ via convolution:
\vspace{-0.1cm}
\begin{equation}
    (x^*,y^*,\theta^*) = \underset{x,y,\theta \in \mathcal{X,Y},\Theta}{\arg\max} \;\boldsymbol{I}_{L}(\theta) \circledast \boldsymbol{I}_{L_{gen}}(x,y)
\end{equation}

\begin{table}[h]
 \footnotesize
 \begin{tabular}{| c || c c c |}
 \hline
 \rule{0pt}{2ex}
  & $e_x$ (pix)$\downarrow$ & $e_y$ (pix)$\downarrow$ &$e_\theta$ (\textdegree)$\downarrow$\\ [0.3ex] 
 \hline\hline
 \rule{0pt}{2.6ex} 
 \underline{\textit{KITTI}} &&&\\
 \rule{0pt}{2.6ex}
 ECML-CT (ours) & \textit{\textbf{4.7}}   & \textit{\textbf{8.2}} & \textit{\textbf{2.12}} \\
 ECML-CNN (ours) & 12.2 & 14.8 & 3.53 \\
 RSL \cite{tang-rslnet} & 22.2 & 23.9 & 4.18 \\
 Canny & 30.4 &	29.8 & 5.21 \\
 Siamese \cite{downes2022siamese} & 31.2 & 34.4  & 6.56 \\
 Classifier & 36.9 & 38.8 & 6.77 \\
 Random & 43.0  & 43.0  & 7.00 \\

 \hline
 \rule{0pt}{2.6ex} 
 \underline{\textit{PandaSet}} &&&\\
 \rule{0pt}{2.6ex} 
 ECML-CT (ours) & \textit{\textbf{8.8}}    & \textit{\textbf{9.3}} & \textit{\textbf{1.70}} \\
 ECML-CNN (ours) & 9.4 & 10.6 & 2.44 \\
 RSL \cite{tang-rslnet} & 30.0 & 32.9 & 3.27 \\
 Canny & 27.2 & 23.2 & 4.80\\
 Siamese \cite{downes2022siamese} & 32.6 & 30.0 & 5.63 \\
 Classifier & 40.7 & 42.3 & 6.64 \\
 Random & 43.0 & 43.0 & 7.00 \\
 
 \hline
 \rule{0pt}{2.6ex} 
 \underline{\textit{Custom A}} &&&\\
 \rule{0pt}{2.6ex} 
 ECML-CT (ours) & \textit{\textbf{3.1}}    & \textit{\textbf{1.6}} & \textit{\textbf{1.69}} \\
 ECML-CNN (ours) & 8.0 & 6.4 & 2.58 \\
 RSL \cite{tang-rslnet} & 29.5  & 29.8 & 1.77\\
 Canny & 19.4 & 16.4  & 4.82\\
 Siamese \cite{downes2022siamese} & 31.0 & 27.1 & 5.76 \\
 Classifier & 39.2 & 44.0 & 6.06\\
 Random & 43.0 & 43.0 & 7.00 \\
 \hline
 
 \rule{0pt}{2.6ex} 
 \underline{\textit{Custom B}} &&&\\
 \rule{0pt}{2.6ex} 
 ECML-CT (ours) & \textit{\textbf{9.7}}    & \textit{\textbf{5.9}} & \textit{\textbf{2.46}} \\
 ECML-CNN (ours) & 28.4 & 24.3 & 5.85 \\
 RSL \cite{tang-rslnet} & 39.8  & 30.8 & 4.96\\
 Canny & 15.3 & 12.3  & 3.89\\
 Siamese \cite{downes2022siamese} & 39.9 & 22.4 & 6.18 \\
 Classifier & 36.1 & 45.5 & 5.98\\
 Random & 43.0 & 43.0 & 7.00 \\
 \hline
\end{tabular}
{
  \caption{\label{tab:comparison}Comparison of methods for mean distance error for x and y, and for mean angle error for $\theta$ on KITTI, Pandaset, Custom A, Custom B. Scale factor is 1.83 m/pixel.}
  \label{tab:results}
}
\end{table}

\begin{figure}[h]
  \includegraphics[width=0.82\linewidth]{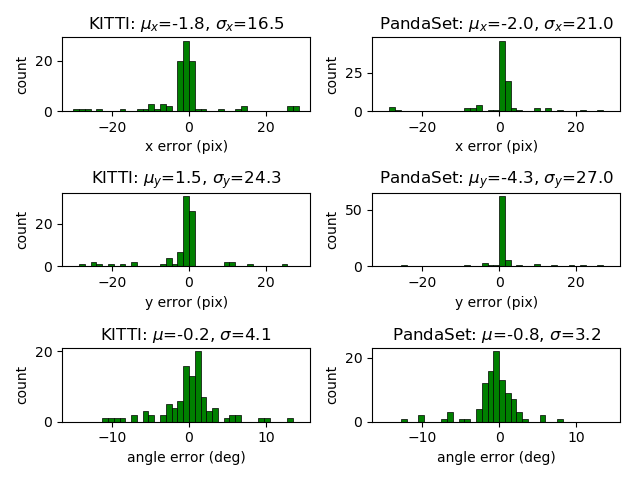}
  \includegraphics[width=0.82\linewidth]{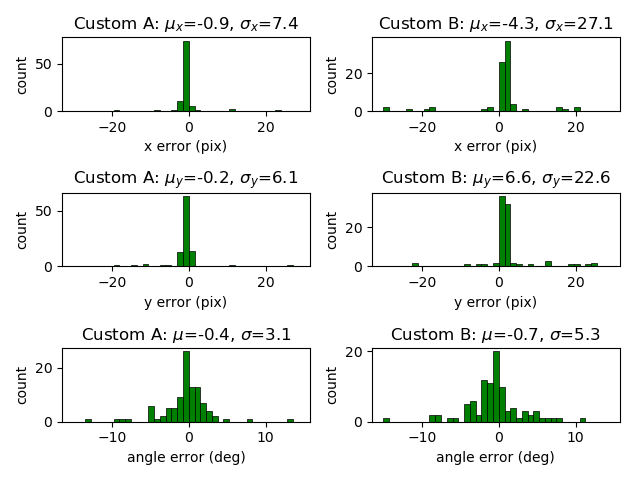}
  {\caption{Distribution of 100 samples for ECML-CT on KITTI, PandaSet, Custom A, and Custom B in x, y, and $\theta$, along with their mean and standard deviation.}
  \label{fig:hist_ct}
  }
\end{figure}

In addition to \cite{tang-rslnet}, we include ECML-CNN as a baseline as well as methods using Canny edges \cite{canny}, metric learning using Siamese networks \cite{downes2022siamese}, and a naive RESNET-based classifier \cite{he2015deep}. Instead of using street-level camera images \cite{downes2022siamese}, we used BEV lidar images as a Siamese network input to correlate with satellite images. 
We show examples of query lidar image and satellite map localization pairs in Fig. \ref{fig:query_map}. Table~\ref{tab:results} summarizes the localization results of our experiment showing our EBM achieving superior accuracy compared to other approaches. 
In Fig.~\ref{fig:hist_ct}, histograms are shown for sample error mean and standard deviation for each dataset using ECML-CT.

\section{Conclusion} \label{conclusion}
We present a new method using an Energy-based Model (EBM) for cross-modal localization between ground range sensors and overhead imagery in GPS-denied environments. Specifically, we demonstrate our approach by leveraging convolutional transformers (ECML-CT) to localize lidar BEV images against satellite maps. 
Although satellite imagery is much easier to obtain than laboriously collected custom lidar maps, the drastic difference in appearance across modalities makes this a very challenging task. 
Our model learns to assess the quality of lidar-satellite image pairs through an efficient training mechanism in order to provide accurate pose estimates for map sizes much greater than those in previous works for cross-modal localization. We perform real-time inference with a two-stage approach
and demonstrate the superior performance of our model in diverse settings.

\clearpage
\bibliographystyle{IEEEtran}

\addtolength{\textheight}{-12cm}   


\end{document}